\documentclass{article}
\usepackage{spconf,amsmath,graphicx}
\usepackage{csquotes}
\usepackage{rotating}
\usepackage{multirow}

\usepackage{times}
\usepackage{epsfig}
\usepackage{subfigure}
\usepackage{color}
\usepackage[normalem]{ulem}

\def\L{{\cal L}}

\title{Salient Object Detection via Bounding-box Supervision}
\name{Mengqi He$^{1}$ \qquad Jing Zhang$^{1}$ \qquad Wenxin Yu$^{2}$}
\address{$^{1}$ Australian National University, Australia, $^{2}$ Southwest University of Science and Technology, China}%
\begin{document}
%
\maketitle
\begin{abstract}
The success of fully supervised saliency detection models depends on a large number of pixel-wise labeling. In this paper, we work on bounding-box based weakly-supervised saliency detection to relief the labeling effort. Given the bounding box annotation, we observe that pixels inside the bounding box may contains extensive labeling noise. However, as the large amount of background is excluded, the foreground bounding box region contains less complex background, making it possible to perform handcrafted features based saliency detection with only the cropped foreground region. As the conventional handcrafted features are not representative enough, leading to noisy saliency maps, we further introduce structure-aware self-supervised loss to regularize the structure of the prediction. Further, we claim that pixels outside bounding box should be background, thus partial cross-entropy loss function can be used to accurately localize the accurate background region. Experimental results on six benchmark RGB saliency dataset illustrate effectiveness of our model.


\end{abstract}
\section{Introduction}
\label{sec:intro}

Salient object detection aims to localize the full scope of the salient foreground, which is usually defined as a binary segmentation task. Most of the conventional techniques are fully-supervised \cite{F3Net_aaai20,wang2020progressive,CVPR2020_LDF}, where the pixel-wise annotations are needed as supervision to train a mapping from input image space to output saliency space. We find that the strong dependency of pixel-wise labeling pose both efficiency and budget challenges for existing fully-supervised saliency detection techniques. To relief the labeling effort, some un/weakly-supervised learning based salient object detection techniques \cite{jing2020weakly,imagesaliency,deepups,Zhang_2018_CVPR,zhang2020learning_eccv,Zhang_2017_ICCV,Guanbin_weaksalAAAI,zeng2019multi,weak_saliency_bounding_box,structure_consistency_scribble,zhang2020learning_tpami} are introduced, which aim to learn saliency from easy-to-obtain labeling, including image-level labels \cite{imagesaliency}, scribbles \cite{jing2020weakly,structure_consistency_scribble}, bounding box supervisions \cite{weak_saliency_bounding_box} or noisy labeling \cite{Zhang_2017_ICCV,Zhang_2018_CVPR,deepups,zhang2020learning_eccv,zhang2020learning_tpami}. In this paper, we present a new bounding box supervision based weakly-supervised saliency detection model.

\begin{figure}[thp]
   \begin{center}
   \begin{tabular}{{c@{ } c@{ } c@{ } }}
   {\includegraphics[width=0.30\linewidth]{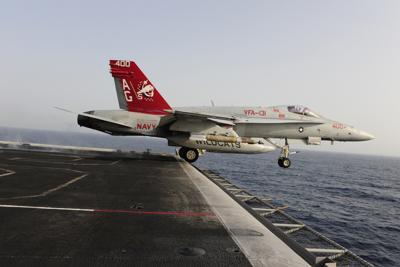}}&
   {\includegraphics[width=0.30\linewidth]{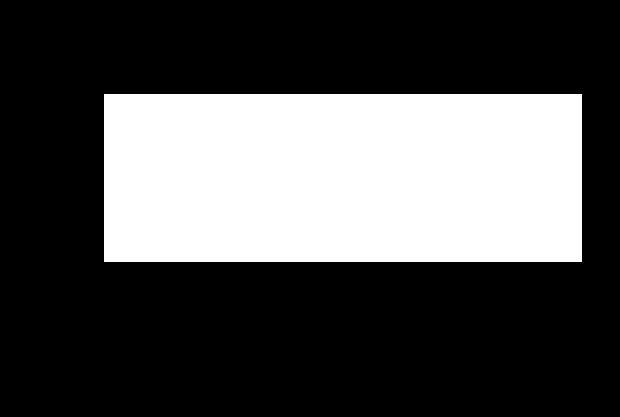}}&
   {\includegraphics[width=0.30\linewidth]{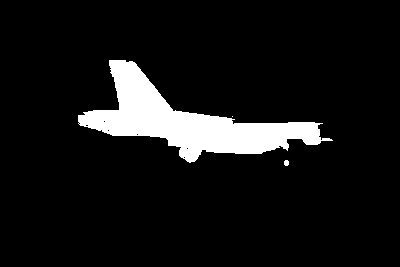}}
   \\
   \footnotesize{Image}&\footnotesize{bndBox}&\footnotesize{GrabCut}\\
   \end{tabular}
   \end{center}
    \caption{Given image (\enquote{Image}) and it's bounding box annotation (\enquote{bndBox}), GrabCut leads to an pseudo segmentation result (\enquote{GrabCut}), serving as pseudo label for the proposed weakly supervised saliency detection model.
    } 
    \label{pseudo_label_generation}
\end{figure}

The advantage of bounding-box supervision is that the background of bounding-box annotation is pure background. Although there still exists background region within the bounding-box foreground annotation, the significantly reduced background region makes it possible to perform conventional handcrafted feature based saliency detection methods on the cropped foreground region. In Fig.~\ref{pseudo_label_generation}, we show the generated pseudo label by combining bounding-box annotation with the conventional handcrafted feature based method, which clearly show it's superiority.

As a deep neural network can fit any types of noise \cite{on_calibration}, directly using the generated pseudo label as ground truth will lead to biased model, over-fitting on the noisy pseudo label. To further constrain the structure of the prediction, we present structure-aware self-supervised loss function. The main goal of this loss function is constrain the prediction with structure well-aligned with the input image. In this way, we aim to obtain structure accurate predictions (see Fig.~\ref{visual_comparison}).

\section{Related Work}
\label{sec:related_work}
\noindent\textbf{Fully-supervised Saliency models:}
The main focus of fully-supervised saliency detection models is to achieve effective feature aggregation
\cite{chen2020global} \cite{zhang2018progressive} \cite{pang2020multi}. Due to the use of stride operation, the resulting saliency maps are usually with low resolution. To produce structure accurate saliency prediction,
some method use edge supervision to learn more feature about the object boundary to refine the saliency predictions using better structure of the object \cite{yu2021structure} \cite{wang2019salient} \cite{zhao2019egnet}.

\begin{figure}[ht!]
   \begin{center}
   \begin{tabular}{{c@{ }}}
   {\includegraphics[width=0.9\linewidth]{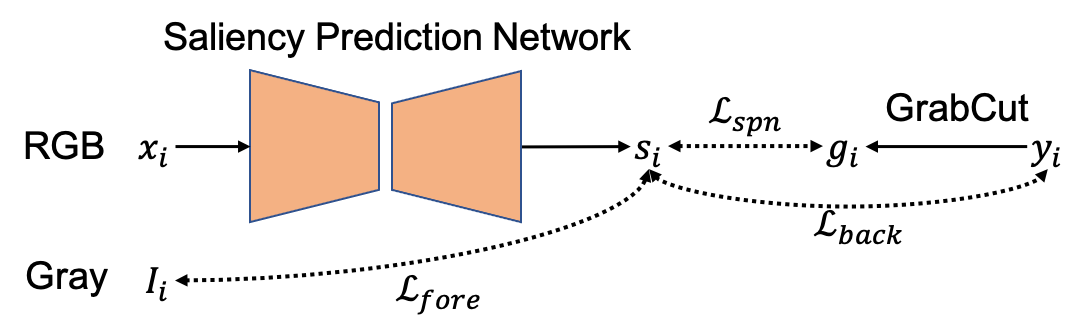}}\\
   \end{tabular}
   \end{center}
    \caption{Given bounding box annotation $y_i$, we first generate pseudo saliency maps $g_i$ via GrabCut. Then, we design saliency prediction network(SPN) to regress the saliency map with the above pseudo saliency maps as supervision ($\mathcal{L}_{spn}$). Due to labeling noise, we introduce a foreground structure-aware loss $\mathcal{L}_{fore}$, namely the smoothness loss \cite{wang2018occlusion} to further constrain the structure of the salient foreground (regions within the bounding box area). As the background of the pseudo saliency map is accurate, we introduce background loss $\mathcal{L}_{back}$ using the partial cross-entropy loss to constrain model prediction in the background bounding box area.
    } 
    \label{overview_pipeline}
\end{figure}

\noindent\textbf{Weakly-supervised Saliency models:}
The goal of weakly-supervised saliency detection is to relief the labeling effort \cite{jing2020weakly,structure_consistency_scribble,imagesaliency}. Given bounding box supervision,
They use mask to create the persudo mask,such as GrabCut \cite{rother2004grabcut}.
\cite{hsu2019weakly} use the crossing line of the bounding box as supervision prior.

\section{Our Method}
\label{sec:our_method}

\subsection{Overview}
In this paper, we study weakly-supervised saliency detection with bounding-box supervision. Specifically, given bounding-box supervision of the training dataset, model is trained to accurately produce saliency maps for testing images. 
Let's define the training data set as: $D = \{x_i,y_i\}_{i=1}^N$ of size $N$, where $x_i$ and $y_i$ are the input RGB image and it's corresponding bounding box supervision, and $i$ indexes the images.
To generate $y_i$, each salient non-overlap salient instance is annotated with separated bounding box, and we generate one single bounding box for the overlapped salient instances.
In this way, the foreground of $y_i$ contains different level of noise depending on the position and shape of the bounding box, and the background of $y_i$ is accurate background.

Four main steps are included in our method: (1) a persudo saliency map generator which use the Grabcut to generate pseudo saliency maps given bounding box supervision;
(2) a saliency prediction network (SPN) to produce a saliency map, which is supervised with above pseudo saliency maps;
(3) structure-aware loss to optimize the foreground predictions;
(4) partial-cross entropy based background loss to optimize the background prediction.

\subsection{Persudo Saliency Map Generator via GrabCut}
Given bounding box supervision, we first generate pseudo saliency maps with Grabcut (see Fig.~\ref{pseudo_label_generation}). Compared with the direct bounding box supervision $y_i$, the generated pseudo saliency map $g_i$ is more accurate in structure, making it suitable to serve as pseudo label for saliency prediction. Specifically, we repeat the Grabcut operation several times until we obtain a reasonably accurate pseudo label.


\subsection{Saliency Prediction Network}
Following the conventional fully-supervised saliency detection, we design a \enquote{saliency prediction network} to generate saliency maps. Specifically, we take the ResNet50 as backbone. Given the backbone feature $f_{\theta_1}(x)=\{s_k\}_{k=1}^4$ ($\theta_1$ is parameters of the encoder (the backbone model), the saliency prediction network aims to generate saliency map $s=f_\theta(x)$, where $\theta=\{\theta_1,\theta_2\}$ with $\theta_2$ as parameters of the decoder. Specifically, we feed each $s_k$ to a $3\times3$ convolutional layer to generate new backbone feature $\{s'_k\}_{k=1}^4$ of same channel size $C=64$. Then, we adopt decoder from \cite{Ranftl2020}, which takes $\{s'_k\}_{k=1}^4$ as input and generate $s=f_{\theta_2}(\{s'_k\}_{k=1}^4)$. Note that, we define parameters of the above 4 copies of $3\times3$ convolutional layers as part of parameters within $\theta_1$.


Given saliency prediction $s_i$, we first design task related loss function, aiming to regress the pseudo salienncy map $g_i$ from GrabCut. Specifically, we adopt the symmetric cross-entropy loss \cite{wang2019symmetric}, which is proven relative robust to labeling noise:
\begin{equation}
    \label{prediction_loss}
    \mathcal{L}_{spn} = \alpha \mathcal{L}_{ce}(s,g) + \beta \mathcal{L}_{ce}(g,s),
    \end{equation}
where $\mathcal{L}_{ce}$ is the cross-entropy loss, $\alpha$ and $\beta$ are used to weigh the contribution of each loss, and empirically we set $\alpha=\beta=1$ in this paper. $s$ and $g$ are the model prediction and pseudo saliency map with GrabCut.
    

\subsection{Foreground Structure Constrain}
Although GrabCut can generate relatively better pseudo label compared with the original bounding box supervision, the complex saliency foreground leads to noisy supervision after GrabCut. To constrain the prediction within the foreground bounding box region, we adopt smoothness loss \cite{wang2018occlusion} to produce structure accurate prediction.

The smoothness loss is defined as:
\begin{equation}
    \label{smoothness_loss}
    \mathcal{L}_{fore} = \sum_{u,v} \sum_{d\in{\overrightarrow{x},\overrightarrow{y}}} \Psi(|\partial_d s_{u,v}|e^{-\alpha |\partial_d (y\cdot I_{u,v})|}),
\end{equation}
where $\Psi$ is defined as $\Psi(s) = \sqrt{s^2+1e^{-6}}$ to avoid calculating the square root of zero, $I_{u,v}$ is the image intensity value at pixel $(u,v)$, $d$ indicates the partial derivatives on the $\overrightarrow{x}$ and $\overrightarrow{y}$ directions. Different with the conventional smoothness loss in \cite{wang2018occlusion}, we introduce gate $y$ to the calculation of smoothness loss to pay attention to the bounding box foreground region inspired by \cite{jing2020weakly}.



\begin{table*}[t!]
  \centering
  \scriptsize
  \renewcommand{\arraystretch}{1.2}
  \renewcommand{\tabcolsep}{0.5mm}
  \caption{\footnotesize{Performance comparison with benchmark saliency detection models.}}
  \begin{tabular}{r|cccc|cccc|cccc|cccc|cccc|cccc}
  \hline
  &\multicolumn{4}{c|}{DUTS \cite{imagesaliency}}&\multicolumn{4}{c|}{ECSSD \cite{yan2013hierarchical}}&\multicolumn{4}{c|}{DUT \cite{Manifold-Ranking:CVPR-2013}}&\multicolumn{4}{c|}{HKU-IS \cite{li2015visual}}&\multicolumn{4}{c|}{PASCAL-S \cite{pascal_s_dataset}}&\multicolumn{4}{c}{SOD \cite{sod_dataset}} \\
    Method & $S_{\alpha}\uparrow$&$F_{\beta}\uparrow$&$E_{\xi}\uparrow$&$\mathcal{M}\downarrow$& $S_{\alpha}\uparrow$&$F_{\beta}\uparrow$&$E_{\xi}\uparrow$&$\mathcal{M}\downarrow$& $S_{\alpha}\uparrow$&$F_{\beta}\uparrow$&$E_{\xi}\uparrow$&$\mathcal{M}\downarrow$& $S_{\alpha}\uparrow$&$F_{\beta}\uparrow$&$E_{\xi}\uparrow$&$\mathcal{M}\downarrow$& $S_{\alpha}\uparrow$&$F_{\beta}\uparrow$&$E_{\xi}\uparrow$&$\mathcal{M}\downarrow$& $S_{\alpha}\uparrow$&$F_{\beta}\uparrow$&$E_{\xi}\uparrow$&$\mathcal{M}\downarrow$ \\ \hline
   SCRN \cite{scrn_sal} & .885 & .833 & .900 & .040 & .920 & .910 & .933 & .041 & .837 & .749 & .847 & .056 & .916 & .894 & .935 & .034 & .869 & .833 & .892 & .063 & .817 & .790 & .829 & .087\\ 
   F3Net \cite{wei2020f3net} & .888 & .852 & .920 & .035 & .919 & .921 & .943 & .036 & .839 & .766 & .864 & .053 & .917 & .910 & .952 & .028 & .861 & .835 & .898 & .062 & .824 & .814 & .850 & .077\\
   ITSD \cite{zhou2020interactive} & .886 & .841 & .917 & .039 & .920 & .916 & .943 & .037 & .842 & .767 & .867 & .056 & .921 & .906 & .950 & .030 & .860 & .830 & .894 & .066 & .836 & .829 & .867 & .076\\
    PAKRN \cite{xu2021locate} & .900 & .876 & .935 & .033  & .928 & .930 & .951 & .032 & .853 & .796 & .888 & .050 & .923 & .919 & .955 & .028 & .859 & .856 & .898 & .068  & .833 & .836 & .866 & .074 \\ 
    MSFNet \cite{zhang2021auto} & .877 & .855 & .927 & .034  & .915 & .927 & .951 & .033 & .832 & .772 & .873 & .050 & .909 & .913 & .957 & .027 & .849 & .855 & .900 & .064  & .813 & .822 & .852 & .077  \\ 
    CTDNet\cite{zhao_CTDNet_ACMMM_2021} &.893 &.862 &.928 &.034 &.925 &.928 &.950 &.032 &.844 &.779 &.874 &.052 &.919 &.915 &.954 &.028 &.861 &.856 &.901 &.064 &.829 &.832 &.858 &.074 \\\hline
    VST\cite{liu_ICCV_2021_VST} &.896 &.842 &.918 &.037 &.932 &.911 &.943 &.034 &.850 &.771 &.869 &.058 &.928 &.903 &.950 &.030 &.873 &.832 &.900 &.067 &.854 &.833 &.879 &.065 \\
    GTSOD \cite{zhang2021learning_nips} &.908 &.875 &.942 &.029 &.935 &.935 &.962 &.026 &.858 &.797 &.892 &.051 &.930 &.922 &.964 &.023 &.877 &.855 &.915 &.054 &.860 &.860 &.898 &.061 \\
   \hline
   SSAL~\cite{jing2020weakly} & .803 & .747 & .865 & .062 & .863 & .865 & .908 & .061 & .785 & .702 & .835 & .068 & .865 & .858 & .923 & .047 & .798 & .773 & .854 & .093 & .750 & .743 & .801 & .108\\ 
   WSS~\cite{imagesaliency}  &.748 &.633 &.806 &.100 &.808 &.774 &.801 &.106 &.730 &.590 &.729 &.110 &.822&.773 &.819 &.079 & .701 & .691 & .687 & .187 & .698 & .635 & .687 & .152  \\
   C2S~\cite{xin2018c2s} &.805 &.718 &.845 &.071 &- &- &- &- &.773 &.665 &.810 &.082 &.869 &.837 &.910 &.053 & .784 & .806 & .813 & .130 & .770 & .741 & .799 & .117\\
   SCWS~\cite{structure_consistency_scribble}  &.841 &.818 &.901 &.049 &.879 &.894 &.924 &.051 &.813 &.751 &.856 &.060 &.883 &.892 &.938 &.038 &.821 &.815 &.877 &.078 &.782 &.791 &.833 &.090 \\ \hline
   
    Ours & .796 & .715 & .821 & .070 & .856 & .838 & .873 & .069 & .792 & .695 & .812 & .072 & .855 & .826 & .882 & .057 & .748 & .760 & .772 & .147 & .740 & .704 & .758 & .119\\ \hline
   
  \end{tabular}
  \label{tab:benchmark_rgb_sod}
\end{table*}

\begin{figure*}[thp]
   \begin{center}
   \begin{tabular}{{c@{ } c@{ } c@{ } c@{ } c@{ } c@{ } c@{ }}}
   {\includegraphics[width=0.13\linewidth]{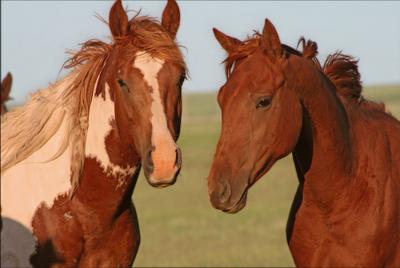}}&
   {\includegraphics[width=0.13\linewidth]{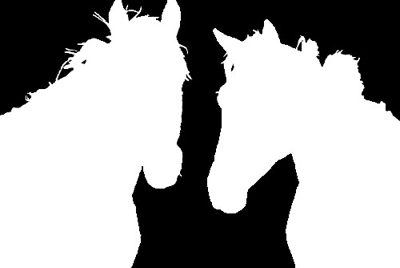}}&
   {\includegraphics[width=0.13\linewidth]{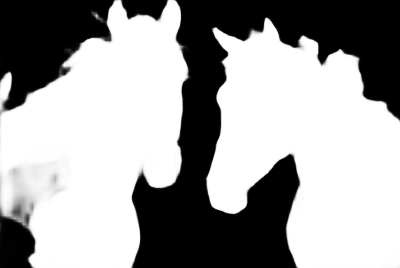}}&
   {\includegraphics[width=0.13\linewidth]{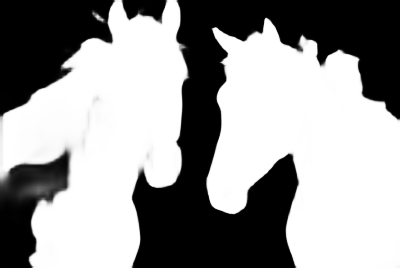}}&
   {\includegraphics[width=0.13\linewidth]{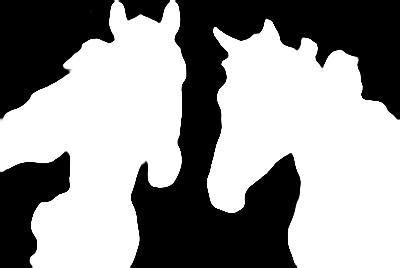}}&
   {\includegraphics[width=0.13\linewidth]{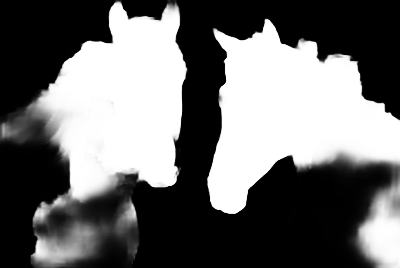}}&
   {\includegraphics[width=0.13\linewidth]{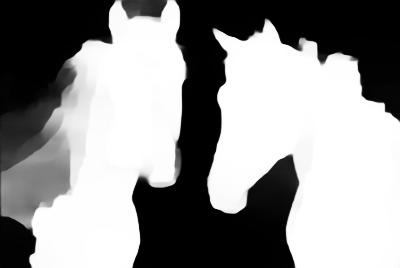}}\\
   \footnotesize{Image}&\footnotesize{GT}&\footnotesize{F3Net \cite{F3Net_aaai20}}&\footnotesize{MSFNet \cite{zhang2021auto}} &\footnotesize{CTDNet\cite{zhao_CTDNet_ACMMM_2021}}&\footnotesize{SSAL~\cite{jing2020weakly}}&\footnotesize{Ours}
   \end{tabular}
   \end{center}
    \caption{Visual comparison with benchmark saliency detection models, where F3Net \cite{F3Net_aaai20}, MSFNet \cite{zhang2021auto} and CTDNet\cite{zhao_CTDNet_ACMMM_2021} are fully-supervised saliency detection models, and SSAL~\cite{jing2020weakly} is a scribble-supervised weakly-supervised saliency detection model.
    } 
    \label{visual_comparison}
\end{figure*}

\subsection{Background Accuracy Constrain}
As background of the bounding box supervision is accurate background for saliency prediction, we adopt partial cross-entropy loss to constrain accuracy of prediction within the background bounding box region. Specifically, given bounding box annotation $y$ and model prediction from the saliency prediction network $s$, we define the background loss as:
\begin{equation}
    \label{background_loss}
    \mathcal{L}_{back}=  \gamma\mathcal{L}_{ce}(s*(1-y),\mathbf{0}),
\end{equation}
where $\gamma=\frac{H*W}{H*W-z}$ ($H$ and $W$ represent image size, $z$ is the number of pixels that are covered in the foreground bounding box), $\mathbf{0}$ is an all-zero matrix of the same size as $s$.


\subsection{Training the Model}
With both the saliency regression loss $\mathcal{L}_{spn}$, foreground structure loss $\mathcal{L}_{fore}$ and background accuracy loss $\mathcal{L}_{back}$, our final loss function is defined as:
\begin{equation}
    \label{final_loss}
    \mathcal{L}_{f} = \mathcal{L}_{spn} + \lambda_1 \mathcal{L}_{fore} + \lambda_2 \mathcal{L}_{back},
\end{equation}
where $\lambda_1$ and $\lambda_2$ is used to control the contribution of the foreground loss and the background loss. Empirically,
we set $\lambda_1=\lambda_2=1$.

We train our final model for 40 epochs. The saliency prediction network is initialized with ResNet50 pre-trained on ImageNet.
The initial learning rate is 2.5e-4, and the decay epoch is 20 while the decay rate is 0.9. As the training batch size be set up to 16 on a PC with an NVIDIA GeForce GTX 1080ti GPU, the training process of the network takes 8 hours.
\section{Experimental Results}
\label{sec:exp_results}

\subsection{Setup}
\noindent\textbf{Data set:}
We train our models using the DUTS training dataset \cite{imagesaliency} $D=\{x_i,y_i\}_{i=1}^N$ of size $N=10,553$, and test on six other widely used datasets: the DUTS testing dataset,  ECSSD \cite{yan2013hierarchical}, DUT \cite{Manifold-Ranking:CVPR-2013}, HKU-IS \cite{li2015visual}, PASCAL-S \cite{pascal_s_dataset} and the
SOD testing dataset \cite{sod_dataset}. The supervision $y_i$ in our case is bounding box annotation.





\begin{table*}[t!]
  \centering
  \scriptsize
  \renewcommand{\arraystretch}{1.2}
  \renewcommand{\tabcolsep}{0.5mm}
  \caption{\footnotesize{Performance of ablation study related experiments.
  }}
  \begin{tabular}{r|cccc|cccc|cccc|cccc|cccc|cccc}
  \hline
  &\multicolumn{4}{c|}{DUTS \cite{imagesaliency}}&\multicolumn{4}{c|}{ECSSD \cite{yan2013hierarchical}}&\multicolumn{4}{c|}{DUT \cite{Manifold-Ranking:CVPR-2013}}&\multicolumn{4}{c|}{HKU-IS \cite{li2015visual}}&\multicolumn{4}{c|}{PASCAL-S \cite{pascal_s_dataset}}&\multicolumn{4}{c}{SOD \cite{sod_dataset}} \\
    Method & $S_{\alpha}\uparrow$&$F_{\beta}\uparrow$&$E_{\xi}\uparrow$&$\mathcal{M}\downarrow$& $S_{\alpha}\uparrow$&$F_{\beta}\uparrow$&$E_{\xi}\uparrow$&$\mathcal{M}\downarrow$& $S_{\alpha}\uparrow$&$F_{\beta}\uparrow$&$E_{\xi}\uparrow$&$\mathcal{M}\downarrow$& $S_{\alpha}\uparrow$&$F_{\beta}\uparrow$&$E_{\xi}\uparrow$&$\mathcal{M}\downarrow$& $S_{\alpha}\uparrow$&$F_{\beta}\uparrow$&$E_{\xi}\uparrow$&$\mathcal{M}\downarrow$& $S_{\alpha}\uparrow$&$F_{\beta}\uparrow$&$E_{\xi}\uparrow$&$\mathcal{M}\downarrow$ \\ \hline
BndBox  &  .642	&  .514	&  .673	&  .167
&  .683	&  .620	&  .712	&  .185
&  .671	&  .533	&  .701	&  .149
&  .683	&  .597	&  .722 &  .167
&  .621	&  .608	&  .669	&  .232
&  .625	&  .552	&  .669	&  .203 \\
GCut  & .795 & .707 &	.814 &	.071 &
.856 &	.834 & .871 & .069 &
.787 &	.681 & .800 & .075 &
.852 &	.820 & .874 & .059 &
.745 &	.751 & .762 & .151 &
.733 &  .689 & .747 & .120\\
FGCut & .787 &  .715 &  .823 &  .069
&  .854	&  .846	&  .883	&  .068
&  .778	&  .687	&  .806	&  .070
&  .851	&  .835	&  .892	&  .055
&  .733	&  .749	&  .760	&  .157
&  .724	&  .694	&  .750	&  .124 \\\hline

Ours   & .796 & .715 & .821 & .070 & 
.856 & .838 & .873 & .069 & 
.792 & .695 & .812 & .072 & 
.855 & .826 & .882 & .057 & 
.748 & .760 & .772 & .147 & 
.740 & .704 & .758 & .119 \\   \hline
  \end{tabular}
  \label{tab:ablation_study_experiments}
\end{table*}

\noindent\textbf{Evaluation Metrics:} Four evaluation metrics are used, including Mean Absolute Error (MAE $\mathcal{M}$), Mean F-measure ($F_{\beta}$), mean E-measure ($E_{\xi}$) and the S-measure $S_{\alpha}$ \cite{fan2018enhanced}





\subsection{Performance Comparison}
\noindent\textbf{Quantitative comparison:} We show performance of our model in Table \ref{tab:benchmark_rgb_sod}, where models in the top two blocks are fully supervised models (models in the middle block are transformer \cite{transformer_raw} based), and models in the last block (not \enquote{Ours}) are weakly supervised models. Performance comparison show competitive performance of our model, leading to an alternative weakly supervised saliency detection model.

\noindent\textbf{Qualitative comparison:} We compare predictions of our model with four benchmark models and show results in
Fig.~\ref{visual_comparison}, which further explain that with both the foreground and background constrains, our weakly supervised model can obtain relative structure accurate predictions. 

\begin{figure}[thp]
   \begin{center}
   \begin{tabular}{{c@{ } c@{ } c@{ } c@{ } c@{ } c@{ }}}

   {\includegraphics[width=0.153\linewidth]{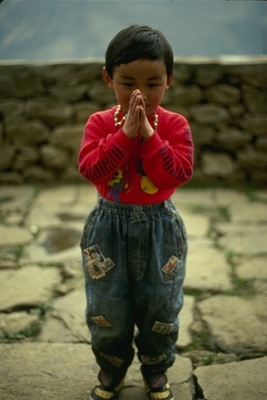}}&
   {\includegraphics[width=0.153\linewidth]{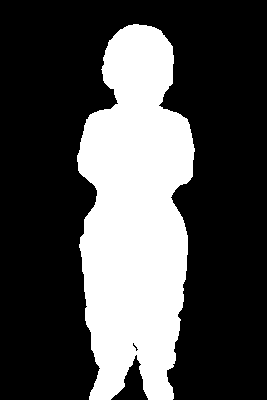}}&
   {\includegraphics[width=0.153\linewidth]{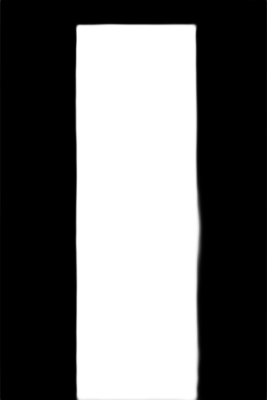}}&
   {\includegraphics[width=0.153\linewidth]{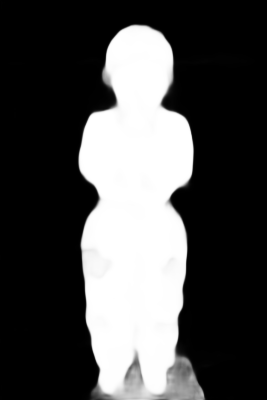}}&
   {\includegraphics[width=0.153\linewidth]{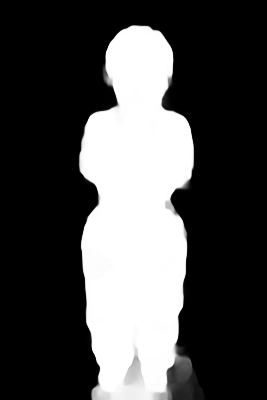}}&
   {\includegraphics[width=0.153\linewidth]{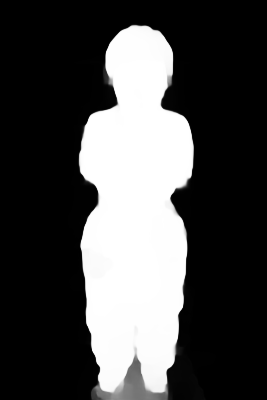}}\\
   {\includegraphics[width=0.153\linewidth]{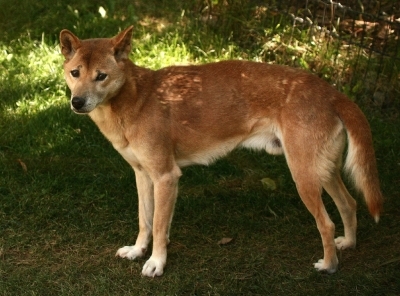}}&
   {\includegraphics[width=0.153\linewidth]{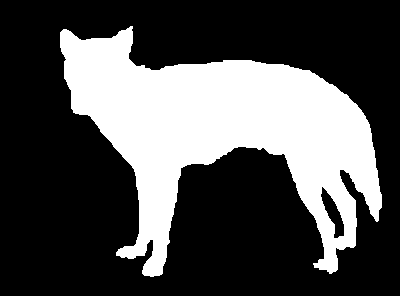}}&
   {\includegraphics[width=0.153\linewidth]{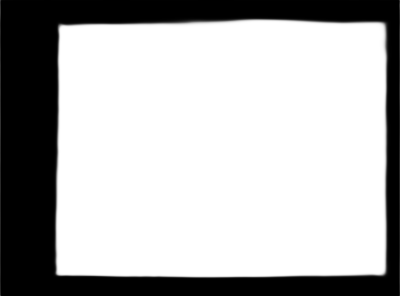}}&
   {\includegraphics[width=0.153\linewidth]{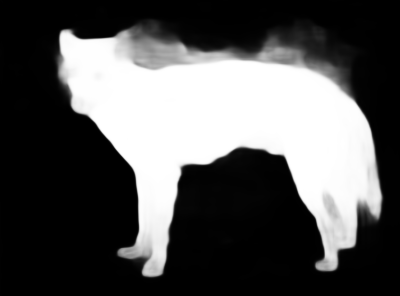}}&
   {\includegraphics[width=0.153\linewidth]{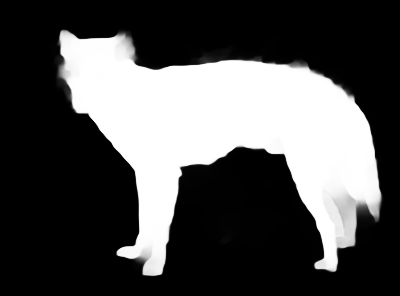}}&
   {\includegraphics[width=0.153\linewidth]{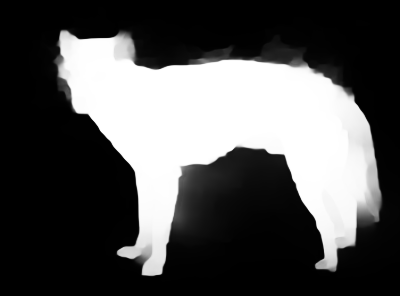}}\\
   {\includegraphics[width=0.153\linewidth]{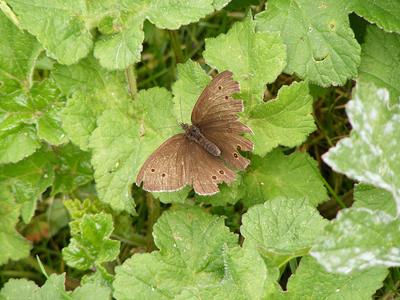}}&
   {\includegraphics[width=0.153\linewidth]{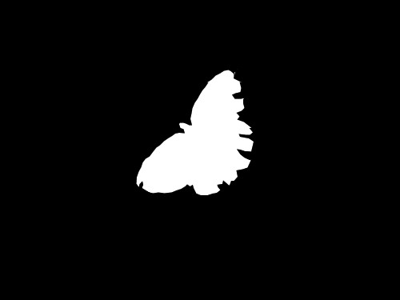}}&
   {\includegraphics[width=0.153\linewidth]{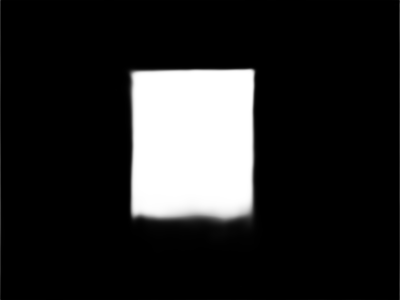}}&
   {\includegraphics[width=0.153\linewidth]{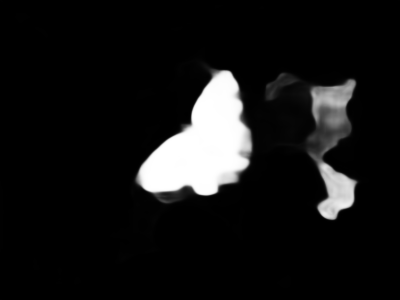}}&
   {\includegraphics[width=0.153\linewidth]{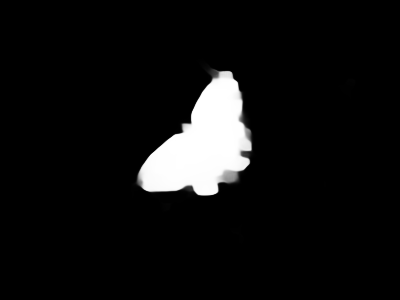}}&
   {\includegraphics[width=0.153\linewidth]{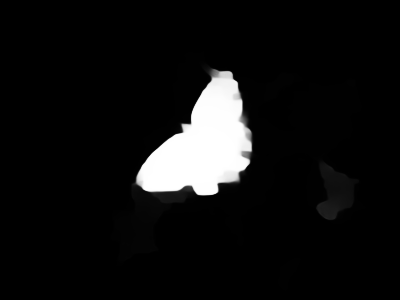}}\\
   \footnotesize{Image}&\footnotesize{GT}&\footnotesize{BndBox}&\footnotesize{GCut} &\footnotesize{FGCut}&\footnotesize{Ours}
   \end{tabular}
   \end{center}
    \caption{Visual comparison of the ablation study related experiments, where each model (\enquote{BndBox}, \enquote{GCut} and \enquote{FGCut}) is introduced in the ablation study section.
    } 
    \label{ablation_visual_comparison}
\end{figure}

\noindent\textbf{Running time comparison:} As both the foreground and background related loss are only used during training. At test time, we can produce saliency maps with the saliency prediction network, leading to an average inference time of 0.02s/image, which is comparable with existing techniques.

\subsection{Ablation Study}

We conducted further experiments to explain the contribution of each component of the proposed model, and show performance of the related experiments in Table \ref{tab:ablation_study_experiments}.
%

\noindent\textbf{Training directly with bounding box supervision $y_i$:} Given bounding box supervision, a straight-forward solution is training directly with the binary bounding box as supervision. We show it's performance as \enquote{BndBox}. 

\noindent\textbf{Training directly with pseudo label from GrabCut:} With the refined pseudo label $g_i$ using GrabCut, we can train another model with $g_i$ as pseudo label directly. The performance is shown as \enquote{GCut}.

\noindent\textbf{Contribution of foreground structure loss:} We further add the foreground structure loss to \enquote{GCut}, leading to \enquote{FGCut}.

\noindent\textbf{Analysis:} As shown in Table \ref{tab:ablation_study_experiments}, directly training with bounding box supervision yields unsatisfactory results, where the model learns to regress the bounding box region (see \enquote{BndBox} in Fig.~\ref{ablation_visual_comparison}). Although the pseudo saliency map with GrabCut is noisy, the model based on it can still generate reasonable predictions (see \enquote{GCut} in both Table \ref{tab:ablation_study_experiments} and Fig.~\ref{ablation_visual_comparison}.). Then, with the proposed foreground and background loss as constrains ($\mathcal{L}_{fore}$ and $\mathcal{L}_{back}$), we obtain better performance with more accurte structures.

\subsection{Model Analysis}
We analyse the model further to explain the advantage and limitations of the proposed method.

\noindent\textbf{Impact of the maximum training epoch:}
We set the maximum epoch as 40 in this paper. As proven in existing noisy labeling literature that longer training time is harmful for noisy labeling setting. We further performed experiments with both longer and shorter training epochs, and observed similar conclusion. We will investigate the optimal training epochs for better performance.



\noindent\textbf{Impact of the feat channel $C$ for new backbonf feature generation in the \enquote{saliency prediction network}:}
For dimension reduction, we feed the backbone feature $\{s_k\}_{k=1}^4$ to four different $3\times3$ convolutional layers to generate the new backbone feature $\{s'_k\}_{k=1}^4$ of channel size $C=64$. We find that model performance is influenced with $C$. The larger $C$ leads to better overall performance. However, the size of the model is also significantly enlarged. To achieve trade off between model performance and training/testing time, we set $C=64$. We will investigate the optimal $C$ in the future.

\noindent\textbf{Edge detection as auxiliary task for prediction structure recovery:} \cite{jing2020weakly} introduced auxiliary edge detection module to their weakly supervised learning framework for structure recovery. We have tried the same strategy and observed no significant performance improvement in our setting. As a multi-task learning framework, the convergence rate of each task is especially important for the final performance, and more sophisticated multi-task learning solution is important to fully explore the contribution of auxiliary edge detection for weakly supervised learning.

\section{Conclusion}
\label{sec:conclusion}
In this paper, we introduce a bounding box based weakly supervised saliency detection model.
Due to the different accuracy of foreground and background, we introduce two sets of loss functions to constrain the predictions within the foreground and background bounding box regions. Experimental results explain the superiority of the proposed solution, leading to an alternative for weakly supervised saliency detection. However, we observe that model performance is sensitive to the maximum training epochs. Longer training will lead to over-fitting on noisy labeling, while shorter training may lead to less effective model. More research on optimal training mechanism for noisy label learning should be investigated.


\footnotesize

\bibliographystyle{IEEEbib}

\end{document}